%% file: pointlearning-arxiv.tex
\documentclass[10pt]{article}

\pdfoutput=1

\usepackage{cite}
\usepackage{amsmath,amssymb,amsfonts}
\usepackage{algorithmic}
\usepackage{graphicx}
\usepackage{textcomp}
\usepackage{xcolor}
\usepackage[draft]{todonotes}
\usepackage{subcaption}
\usepackage[numbers]{natbib}
\usepackage{array}
\usepackage{authblk}

\newcommand\blfootnote[1]{%
  \begingroup
  \renewcommand\thefootnote{}\footnote{#1}%
  \addtocounter{footnote}{-1}%
  \endgroup
}

\newcommand{\citea}[1]{\cite{#1},}
\newcommand{\norm}[1]{\left\lVert#1\right\rVert}
\newcommand*\rot{\rotatebox{90}}

\begin{document}

\title{IPC-Net: 3D point-cloud segmentation using deep inter-point convolutional layers}

\author[1]{Felipe Gomez Marulanda}
\author[1]{Pieter Libin}
\author[1]{Timothy Verstraeten}
\author[1]{Ann Now\'{e}}

\affil[1]{Artificial Intelligence Lab, Department of computer science, Vrije Universiteit Brussel, Brussels, Belgium}

\date{}

\maketitle

\begin{abstract}
\blfootnote{Accepted and presented at the International Conference on Tools with Artificial Intelligence (ICTAI 2018).\\“$\copyright$ 2018 IEEE.  Personal use of this material is permitted.  Permission from IEEE must be obtained for all other uses, in any current or future media, including reprinting/republishing this material for advertising or promotional purposes, creating new collective works, for resale or redistribution to servers or lists, or reuse of any copyrighted component of this work in other works.”}
Over the last decade, the demand for better segmentation and classification algorithms in 3D spaces has significantly grown due to the popularity of new 3D sensor technologies and advancements in the field of robotics. Point-clouds are one of the most popular representations to store a digital description of 3D shapes. However,  point-clouds are stored in irregular and unordered structures, which limits the direct use of segmentation algorithms such as Convolutional Neural Networks.  The objective of our work is twofold: First, we aim to provide a full analysis of the PointNet architecture to illustrate which features are being extracted from the point-clouds. Second, to propose a new network architecture called IPC-Net to improve the state-of-the-art point cloud architectures. We show that IPC-Net extracts a larger set of unique features allowing the model to produce more accurate segmentations compared to the PointNet architecture. In general, our approach outperforms PointNet on every family of 3D geometries on which the models were tested. A high generalisation improvement was observed on every 3D shape, especially on the rockets dataset. Our experiments demonstrate that our main contribution, inter-point activation on the network's layers, is essential to accurately segment 3D point-clouds.
  
\end{abstract}

\input{intro}

\input{relatedWork}
\input{experiments}

\section*{Acknowledgment}
Felipe Gomez Marulanda's work was supported by Doctiris-innoviris brussels grant. 
Pieter Libin and Timothy Verstraeten were supported by a PhD grant of the FWO (Fonds Wetenschappelijk Onderzoek-Vlaanderen).

\newpage
\bibliographystyle{plainnat}  % do not change this line!
\bibliography{refs} 

\end{document}

%% file: intro.tex
\section{Introduction}
The ability to directly learn from unordered data (i.e., 3D point clouds or 3D geometrical shapes) remains an open question. An ample amount of research has been done on extracting representations from  ordered structures, so they can be used to achieve classification or segmentation of 3D spaces. Usually, most methodologies condense the 3D representation into geometrical features that summarise the global and local attributes of the shape. Transforming the 3D space often comes with a negative impact on the accuracy of the segmentation or classification task \cite{Su2015}. The majority of  segmentation and object recognition problems rely on state-of-the-art algorithms such as Convolutional Neural Networks (CNN) to exploit the spatial information that exists within the input space of the problem. CNNs are powerful algorithms for object recognition and are known for outperforming human accuracy on several cases \cite{mdy166,russakovsky2015imagenet}. However, CNNs cannot be directly used on the 3D shapes as the convolutions are ill-suited for extracting spatially-local correlations in irregular and unordered data \cite{lecun2015deep}.

We introduce a CNN architecture that exploits the local correlations that exist within neighbouring points in a 3D point cloud to improve the accuracy of predicting segmentations in the 3D space. Due to the popularity across different fields of research (i.e., in robotics and 3D sensors) \cite{sansoni2009state}, a point-cloud representation is convenient as a large proportion of 3D spaces can be represented as a point-cloud. Our research builds upon two prior studies \cite{Dieleman2016,Qi2017} that demonstrated that point clouds can directly be used in a Neural Network that learns to approximate functions that induce invariance towards rigid transformations such as rotations in the point-cloud. In this manuscript we analyse these types of network and show that the architecture proposed in \cite{Qi2017} ignores the spatial information that exists within clusters of neighbouring points. Furthermore, we propose a solution to this problem in the form of an improved architecture and call it \emph{Inter-Point Convolutions Network} (IPC-Net). In summary, Section \ref{sec:pointNet}  will provide an introduction on the PointNet architecture. Section \ref{sec:pointNetView}  will illustrate how the PointNet kernel activations partitions the 3D space but omits inter-point neighbouring information. Finally, in section \ref{sec:convpointNet} we propose our main contribution which exploits inter-point activations to achieve high segmentation accuracy.

%% file: relatedWork.tex
\section{Related Work}\label{sec:relwork}

Different methods have been proposed to solve classification and segmentation problems for 3D geometries \cite{theologou2015comprehensive}. State-of-the-art techniques translate the geometry into a representation that learning algorithms can understand. This is often achieved by summarising the 3D shapes into geometrical features (i.e., characteristics). In the literature, most feature based methods are divided into 2 categories: \emph{local features methods} (LFM) and \emph{global features methods} (GFM). Both categories represent a geometry in different ways. LFMs target the local characteristics of neighbouring information. For example, computing the curvature in a subset of the input space. In contrast, GFMs characterise the global shape of the geometry by considering the entire input space at once.

Osada et al. developed a method using \emph{shape distributions} \cite{Osada2002}, where the concept of shape functions is designed to measure geometrical characteristics (e.g., functions that calculate distances or angles between arbitrary points). They uniquely identify a 3D geometry by a probability shape distribution generated from one or more shape functions. These unique signatures can be used for classification and shape retrieval problems. Shape distributions is one of the models that belong to the category of GFM, as they summarise the overall geometry in a single distribution. Other methods however attempt to make a trade off between local and global features. Hang Su et al. rendered a collection of images by taking snapshots from different view-angles of 3D geometries \cite{Su2015}. These images are subsequently fed into an ensemble of Convolutional Neural Networks (CNN) to generate \emph{view-based descriptors} (e.g., descriptors generated by images). Hang Su found that view-based descriptors encompass a good balance between local and global features in comparison to  more complex structures such as voxel-based representations\cite{Brock2016}. This is due to the fact that  rendered images are characterised by highly dense representations (pixels) and hence facilitate the extraction of representative 3D features.  

Despite the high accuracy of view-based descriptors, such models still require the transformation of the original 3D format into lower level representations, thereby losing important information. For instance,  view-based descriptors \cite{Su2015} dismiss internal segments of 3D shapes as the rendering only considers the exterior of the geometry. Rendering the internal segments of the geometry would be unfeasible as it requires a combination of affine transformations (i.e., rotation, translation, shear and scaling) to capture the geometry. To counter these concerns, Qi et al. proposed the \emph{PointNet} algorithm that learns directly from 3D meshes and point clouds \cite{Qi2017}. Their learning framework requires no additional transformations, which provides an advantage over the algorithms mentioned. Point-clouds or meshes are more challenging to learn from, compared to other representations such as view-based descriptors. For example, point clouds do not contain an underlying spatial or temporal order, in contrast to pixels in images or samples of a signal. As a result, Neural Networks or other machine learning algorithms cannot be used directly on point clouds. Qi et al proposed an innovative solution to extract global and local geometrical features by approximating \emph{symmetric functions}. A symmetric function is a function that remains unchanged by any permutation of its arguments\cite{WeissteinEW}. In contrast to view-based descriptors, PointNet is sensitive to the density representation of the point cloud. It is common that different point clouds datasets contain regions with non-uniform density areas. This results into a combination of different density sets that may highly decrease the performance of the leaning algorithm. PointNet++ \cite{qi2017pointnet++} circumvents the limitation of different density representation by adapting learning layers to combine features from multiple scales. As our contribution builds upon their Neural Network architecture, we will further describe the PointNet method in Section \ref{sec:pointNet}. 

\section{PointNet}\label{sec:pointNet}

Considering that the IPC-Net builds upon the PoinNet architecture, in the following sections we will further describe the PointNet architecture and provide the necessary background to understand our analysis in Section \ref{sec:pointNetView}.

\subsection{Neural Networks}\label{sec:NN}
Neural Networks (NN) are composed of computational units (i.e., neurons) arranged hierarchically by a set of interconnected layers. Information flows back and forth from the lower to the higher layers of the network allowing it to learn higher order representations of the input data. Intuitively, each computational unit tries to learn a specific characteristic of their input. Connections between input and output units are represented as weights and biases expressing the importance of the respective inputs to the outputs. The objective of NNs is to find the weights and biases which minimise the cost function of the network. The cost function is a way to quantify  the objective of the Neural Network. The most commonly used objective functions are the Mean Square Error\cite{allen1971mean} and Cross-Entropy\cite{shore1980axiomatic} functions. Gradient based functions such as (stochastic) gradient descent, conjugate gradient and Adam \cite{bottou2010large,moller1993scaled,kingma2014adam} are the most popular methods to minimise cost functions in a NN. Finally, a technique called back propagation is used to propagate back the cost of the output layer to every unit in the network. The minimisation phase is an iterative process and it will stop until a desired cost is reached or a condition is fulfilled.

\begin{figure}%[h!]
  \centering
    \includegraphics[width=.6\textwidth]{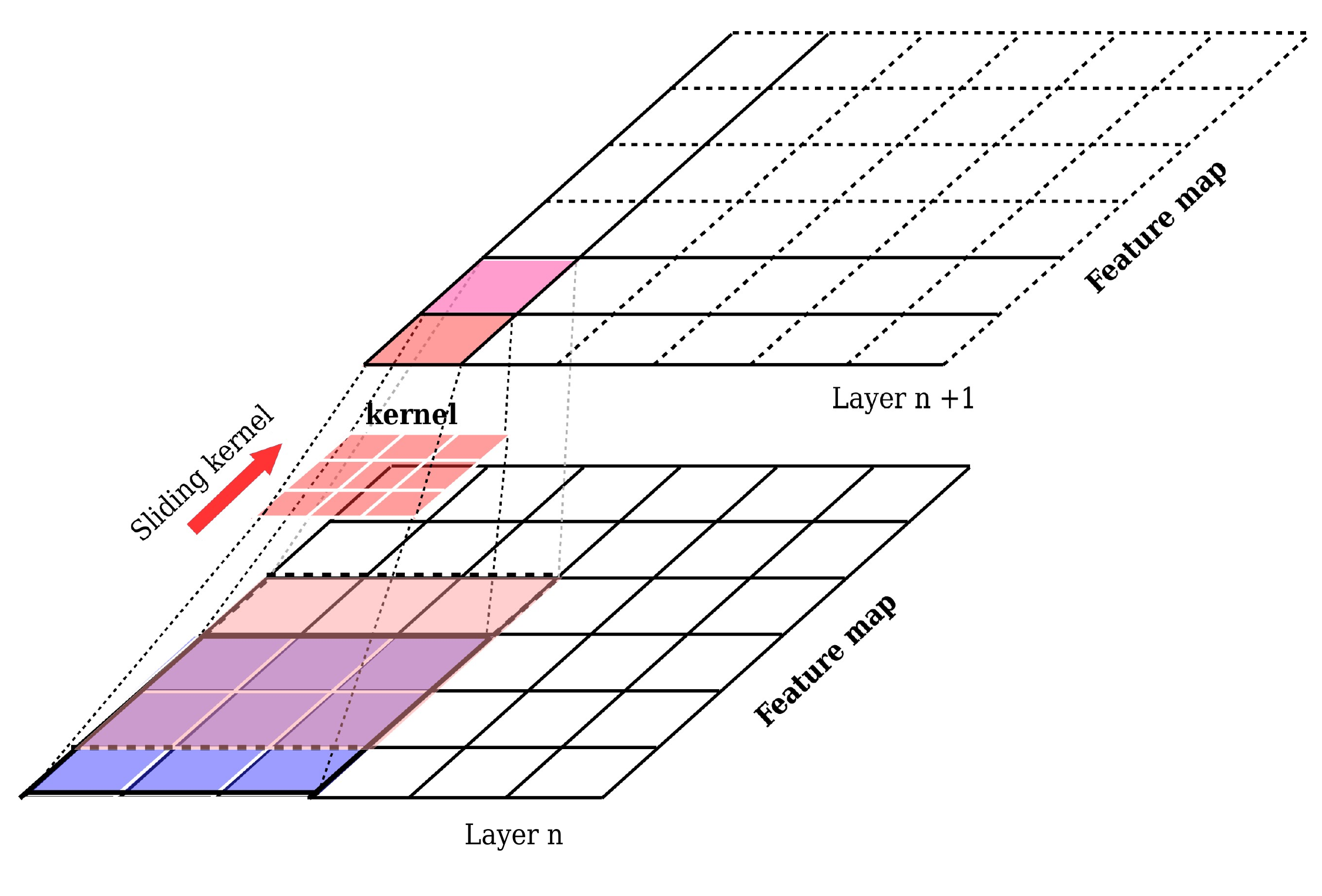}
    \caption{This Figure illustrates the convolution between a 3 by 3 kernel and a 6 by 6 feature map (input). The kernel slides throughout the input space  by element-wise multiplications. The result is summarised by a summation and stored as a scalar in the following feature map. This process is finished until all the input space has been traversed.}
  \label{fig:conv}
\end{figure}

One of the most popular network architectures for image segmentation and classification are the Convolutional Neural Networks (CNN)\cite{krizhevsky2012imagenet}. CNNs can directly learn from  multidimensional arrays (e.g., 2-dimensional images) by introducing 3 new architectural concepts namely: \emph{Local Receptive field}, \emph{shared weights} and \emph{pooling}. A Local Receptive Field allows each layer of the network to introduce a focal view of the input space. This view is called a \emph{receptive field} which is defined as a patch of the input space that a particular CNN's feature is looking at. In a convolutional hidden layer, units are organised as \emph{feature maps} which are the result of convolving a matrix of weights, called \emph{kernels}, with previous feature maps. The convolution in a layer begins by sliding over the input feature map and perform  an element-wise computation of the 2 matrices and sums the results. This process is shown in Figure \ref{fig:conv} 

Every unit in a specific feature map \emph{shares} the same kernel, allowing it to extract patterns that are present across the input space. This dramatically reduces the amount of weights that are needed to train the network. Pooling is performed right after the convolution, where the main idea is to further summarise the features that were captured by the feature maps. This is done by  compressing the information generated by different feature maps either by extracting the maximum or the mean activation of neighbouring units. Doing this removes redundant information encoded within the feature maps and increases the spatial invariance of the input. As a result it makes the model more robust to rigid transformations in the image such as rotation and translation.

\subsection{PointNet architecture}\label{sec:NN}

Convolutional Neural Networks are a perfect choice when dealing with regularly ordered input domains such as images, as CNNs exploit the spatial-local correlations that exists within the pixel representation. Nevertheless, a 3D point cloud is an irregular and unordered representation for which convolutions that leverage spatial correlations are ill-suited. Ideally, the point-cloud could be ordered to exploit the points' spatial information and to extract local and global signatures of the 3D shape. However, some attempts such as \cite{jaderberg2015spatial,vinyals2015order} did not manage to achieve an acceptable accuracy when ordering the inputs. Zaheer et al. \cite{zaheer2017deep}  and Qi, Charles R et al. \cite{Qi2017} proposed to approximate a symmetric function to introduce invariance in the point set. The PointNet architecture compresses the point-cloud into a smaller set of features that roughly corresponds to the skeleton of objects. The algorithm starts by transforming the input space into its canonical representation using a symmetric function. Then it extracts the important features from this representation which results in a new representation of the feature space. This representation can be further aligned by computing an additional affine transformation. Since these transformations have a higher number of dimensions (i.e., 64 x 64) than the input transformations (i.e., 3x3), the feature transformation matrix is constrained to be close to an orthogonal matrix allowing the preservation of its symmetric inner product\cite{Qi2017}. To achieve this, they regularised the cost function by the following equation:
\begin{equation}
    L_{reg} = ||I - AA^T ||^2_F
    \label{eq:norm}
\end{equation}
Where $A$ is the feature transformation matrix approximated by the network, $I$ is the identity matrix and $|| \cdot ||_F$ is the Frobenius norm of a matrix. Qi, Charles et al. found that adding the regularisation term $L_{reg}$ to the cost function stabilises the optimisation. The global features of the input shape are extracted by a \emph{maxpooling layer}, that summarises the activation of each point into a single feature activation. This results into a feature vector that uniquely represents the overall 3D shape. After extracting the local and global features of the 3D shape, these features are aggregated and used to classify or segment the 3D shape.

\begin{figure}%[h!]
  \centering
    \includegraphics[width=0.9\textwidth]{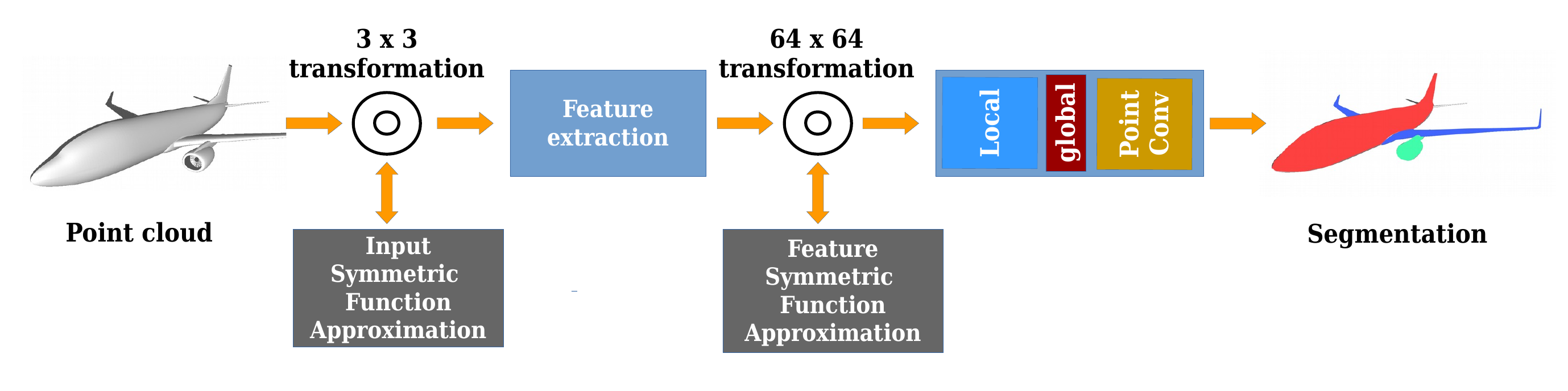}
    \caption{This Figure shows a summary of the original architecture of the PointNet \cite{Qi2017}. Initially, the input shape is passed through the network that is in charge of approximating the symmetric function for the input space (first grey box). Then the original point-cloud is transformed by this symmetric function and passed to the Feature extraction module (second grey box). Similarly to the first phase, the features of the 3D shape are also passed and transformed by another symmetric function. Until this point only the local features for each point have been extracted. A maxpooling layer is used to extract the global signatures of the point-cloud. Finally both global and local features are grouped together to perform a segmentation task.}
  \label{fig:pointnet}
\end{figure}

A summary of the architecture can be observed in Figure \ref{fig:pointnet}. It only illustrates the segmentation path of their architecture as we want to stress the combination of global and local features. The segmentation layer in the PointNet architecture predicts the probability that each point belongs to a particular segmentation by using the aggregated global and local features of the previous layers.

Until now, we described how PointNet uses symmetric functions and maxpooling layers to extract global and local features of the 3D shape. Aggregating these signatures allows the algorithm to achieve high accuracy when segmenting point clouds. However, we postulate that the PointNet architecture still dismisses information that is useful for the segmentation of 3D geometries. A part of this paper (Section \ref{sec:pointNetView}), is devoted to showing how this information is being disregarded by analysing the PointNet architecture's kernel activations of the hidden units.

\section{A new perspective on the PointNet}\label{sec:pointNetView}
In Section \ref{sec:relwork}, we wrote that PointNet extracts the local and global features of 3D point clouds by means of Neural Networks and symmetric function approximations. According to the findings of \cite{Qi2017}, a general function that defines a point set $P = \{p_1, \cdots, p_n\}$ can be approximated by applying symmetric functions on every element in the set as shown in
\begin{equation}
  f(\{p_1,\cdots,p_n\}) \approx g(h(p_1),\cdots,h(p_n))
  \label{eq:symm}
\end{equation}

, where $h$ is a symmetric function modelled by a Neural Network and $g$ is a combination of $h$ and maxpooling functions. Based on several combinations of $g$, different representations of $f$ can be learned. PointNet  aggregates these groups of functions into a single $K$-dimensional features $f_k$ which we called  \emph{kernel features}. They encompass different properties of the set $P$ that are considered to be robust under transformations and generic to a variety of 3D shapes families. In this Section we will provide a different perspective of the $g$ and $h$ functions to further improve the  understanding of the properties that are being extracted from the point set. This is achieved by analysing the activations of a subset of kernel features $f_k$ in the PointNet architecture. Our objective is to employ this analysis to improve the original network architecture. 

Initially, we analysed the kernels that are not part of the symmetric function approximation and inspected the activations of the remaining kernels in the architecture. The kernels of PointNet were activated by introducing a targeted set of 3D shapes and were visualised similarly as kernels for image classification/segmentation \cite{zeiler2014visualizing}. 

\begin{figure}[h!]
  \centering
  \includegraphics{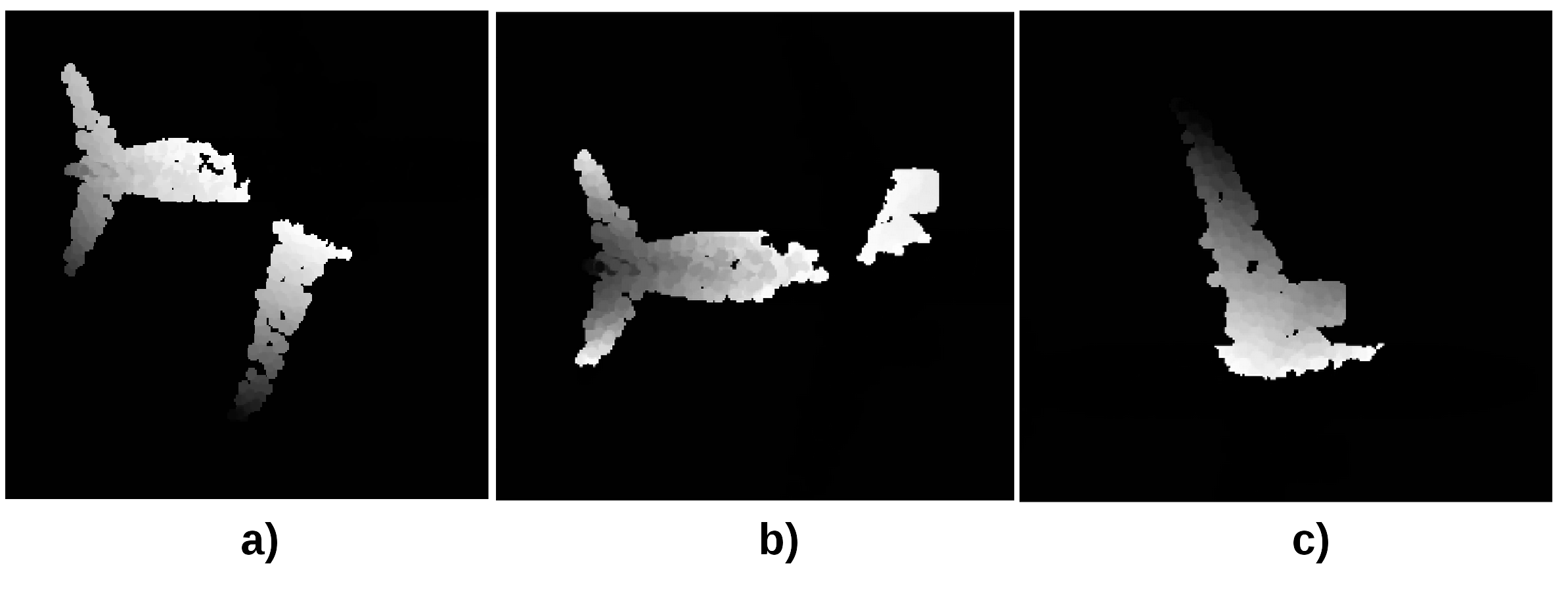}
    \caption{This figure show the 2D field-view of 3 kernels after passing an aircraft geometry through the network. Each of these figures activates a subsection of the 3D space. Only the x and y coordinates of the space are shown in these figures. We illustrate three kernel activations: image a) activates the wing and the tail of the aircraft, image c) shows the activation of the wing and its engine and image b) shows the activation of the engine and the tail. We interpret these activations as a combination of 2D planes that filter the information for a particular view}
  \label{fig:kernel-view}
\end{figure}

In the input and feature kernels group, we found that each kernel $f_k$ is learning a complex combination of 2D planes that partitions the 3D space. This complex combination of planes allows the activation of only one specific part of the space. Our interpretation is aligned with the initial findings of \cite{Qi2017} which states that point features $h(p_i)$  highlight the important local sections of the geometry. However, our interpretation is  more general as it encompasses both  a point feature that emphasise the local signatures and the combination of partitions generated by these kernels. An example of kernel partitions can be seen in Figure \ref{fig:kernel-view}. A complex combination of 2D planes selects a subset of points or features that are relevant to the overall goal of the learning algorithm. This cluster can be either a fundamental composition or compositions of different parts of the 3D shape.

Consequently, we show that the architecture in PointNet is learning to find the optimal partitions of the 3D space that lead to the discovery of the  principal components of the 3D geometry. Furthermore, aggregating these partitions will yield a global shape signature that provides a unique characteristic for every geometry that belongs to the same family of shapes. From this perspective, we can slightly shift the objective of the symmetric function towards learning affine transformation matrices that optimise the partitioning of the 3D space. This does not discard the meaning of the symmetric function mentioned in \cite{Qi2017}. Instead, it adds an extra layer of interpretability. As an example, our analysis showed that the symmetric function will approximate a shear transformation matrix that separates two 3D segmentation surfaces that are close in Euclidean space. Consequently, for the learning algorithm, it becomes easier to find a set of partitions that fragments these two surfaces.

After further analysis of the kernels, we concluded that the PointNet architecture discounts the inner-information that exists within the different partitions of the feature space. For instance, in Figure \ref{fig:kernel-view}c we can observe that most of these points are near each other. This means that we could potentially extract extra features from this ensemble of points. Nonetheless, a convolution cannot be directly applied to these points as they are not spatially ordered in the feature space. Therefore, to exploit the available information on the kernels, we need to spatially group this set of points in order to extract inner-features by means of convolution. In Section \ref{sec:convpointNet} we will describe a new model that uses partition kernels and inner-kernel information to achieve high accuracy for segmentation problems. 
%"order" the  \todo{correct from here}In the 1D space with order data such as images, convolution can be directly implemepemented to extract the spatial-local correlations that exist with the pixels on the image. However, as explained in section \ref{sec:pointNet} convolutions cannot be directly applied on the point clouds as there is not spatial order in the point cloud. However, we just shown in figure \ref{fig:kernel-view} that each kernel provides information of points that are closed to each other or points that may be far from each other but have meaningfull information for the goal of the model. In the following section we will decribe a modify version of the PointNet architecture that exploits the this iner-kernel information.

\section{Inter Point Convolution Network (IPC-Net)}\label{sec:convpointNet}

In this section we will explain an extension of the PointNet architecture that uses Convolutional Neural Networks to exploit the inter-local kernel information. We show that these inner-features build a new set of attributes that exist within points in a field-view of a kernel.

\begin{figure}[h!]
  \centering
  \includegraphics[width=0.9\textwidth]{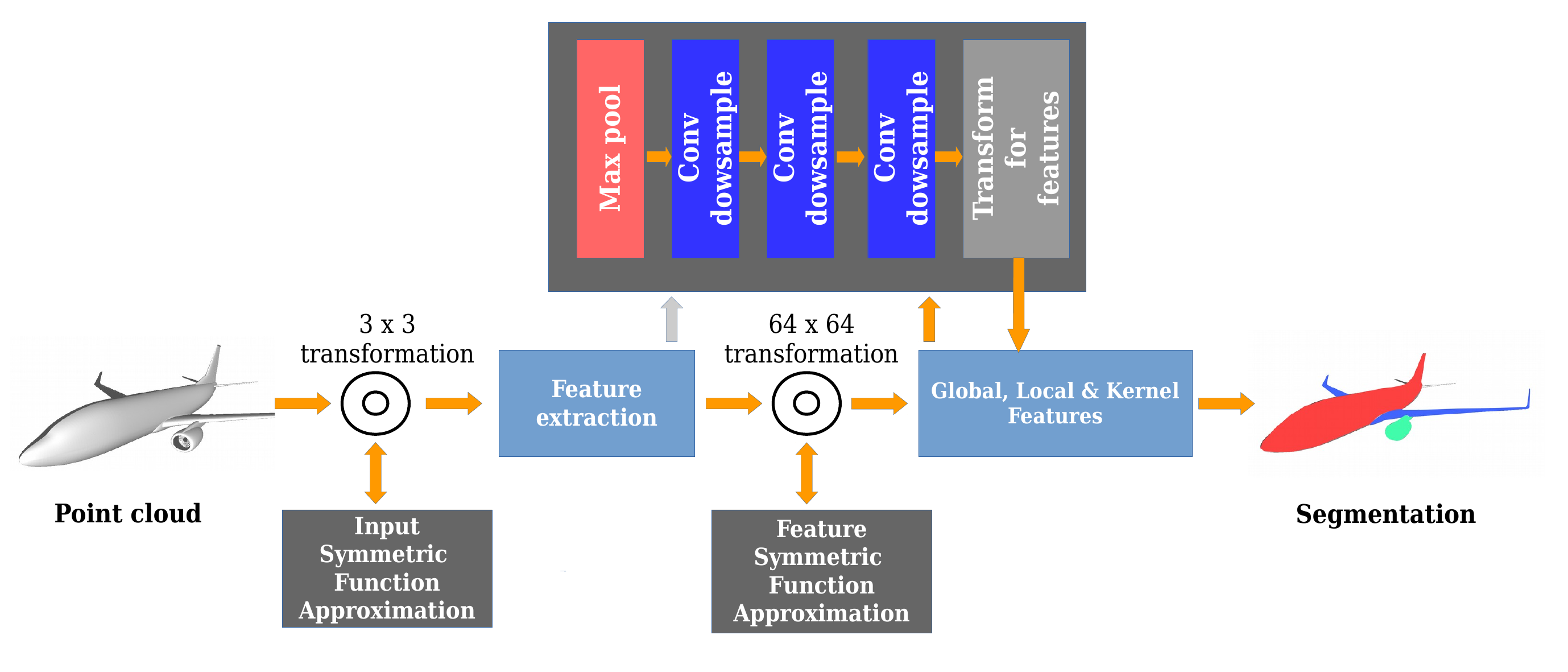}
  \caption{This Figure illustrates the IPC-Net architecture. The kernel activations of the feature transformations layers are given to an external network that is responsible of extracting the neighbouring activation features. This network uses a combination of maxpooling and convolutional layers to extract the inner-features that exist within neighbouring activations. The results are brought back to the initial network to be aggregated with the local and global features.}
  \label{fig:conv-pointNet}
\end{figure}

Similar to the symmetric function, we build an external Convolutional Neural Network to extract the disregarded local features that exist within the kernel activations. In this CNN we initially make use of the maxpooling operation to remove most of the zero values of layers that have been activated by a  kernel. This ensures that the neighbouring points' features in Euclidean space are group together in the activation matrix as shown in figure \ref{fig:norm-kernels}. The resulting set of features is convolved and downsampled by a new set of kernels which encode the neighbouring characteristics of the set. Similar to PointNet, the global signatures are extracted to ensure that the overall geometry is taken into account. We finalise the architecture by aggregating the local, neighbouring and global features into a single feature tensor. This results in a method that embodies a richer set of features compared to those of PointNet. A summary of the inter-point layers shown in Figure \ref{fig:conv-pointNet} is provided in Table \ref{tab:inter-arch}. In Section \ref{sec:exp_rst} we will validate our hypothesis by showing that this newly improved set of features enhances the performance and accuracy of the model for 3D point segmentation.
{
\setlength{\tabcolsep}{0.45em}
\begin{table}[htbp]
  \caption{Illustrates the inter-point layers of the IPC-Net. Each column shows the characteristics of each inter-point layer. The last layer of the table is composed of  reshape and concatenation operations that yield a feature matrix with 1392 channels.}
  \begin{center}
		\scriptsize
		\begin{tabular}{| >{\centering\arraybackslash}m{0.9cm} |>{\centering\arraybackslash}m{1cm}|>{\centering\arraybackslash}m{1cm}|>{\centering\arraybackslash}m{1cm}|>{\centering\arraybackslash}m{0.8cm}|>{\centering\arraybackslash}m{0.8cm}|>{\centering\arraybackslash}m{1cm}|}
			\hline 
			Layers  & Feature extraction & zero removal & down-sample1 & down-sample2 & down-sample3 &transform-concat \\\hline\hline
		  Type   & Conv & maxpool & Conv & Conv & Conv & Concat \\\hline	
		  Channels & 64    & -    & 32   & 16   & 8  	 & 1392 \\\hline	
		  Kernel   & 1x64  & 10x1 & 6x1  & 4x1  & 3x1  &  - \\\hline	
		  Stride   & 1x1   & 10x1 & 5x1  & 3x1  & 2x1  &  - \\\hline			 
		\end{tabular}
                \label{tab:inter-arch}
		%\label{tab:properties}
	\end{center}
\end{table}
}

%% file: experiments.tex
\section{Experiments \& Results}\label{sec:exp_rst}
\begin{figure}%[!thbp]
    \centering
      \mbox{
      \begin{subfigure}{0.5\textwidth}
         \centering
          \includegraphics[scale=0.16,clip, trim=6cm 7.0cm 5cm 3cm]{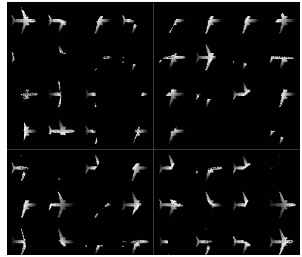}
          \caption{Lower features}
          \label{fig:norm-lkernels}
      \end{subfigure}
      \begin{subfigure}{0.5\textwidth}
        \centering
        \includegraphics[scale=0.16,clip, trim=6cm 7.8cm 5cm 3cm]{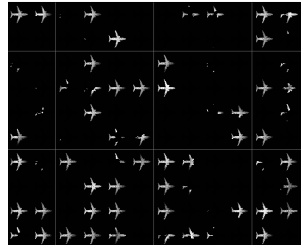}
          \caption{Higher features}
          \label{fig:norm-hkernels}
      \end{subfigure}
      }
      \caption{These figures visualise a 2D grid projection of the kernels of the lower and higher layers in the PointNet architecture. Each image is one kernel feature in the hidden layer and it shows the unit activation when an input is given to the NN.}
      \label{fig:norm-kernels}
\end{figure}

\begin{figure*}%[!htbp]
    \centering
      \mbox{
      \begin{subfigure}{0.5\textwidth}
         \centering
          \includegraphics[ width=1\textwidth]{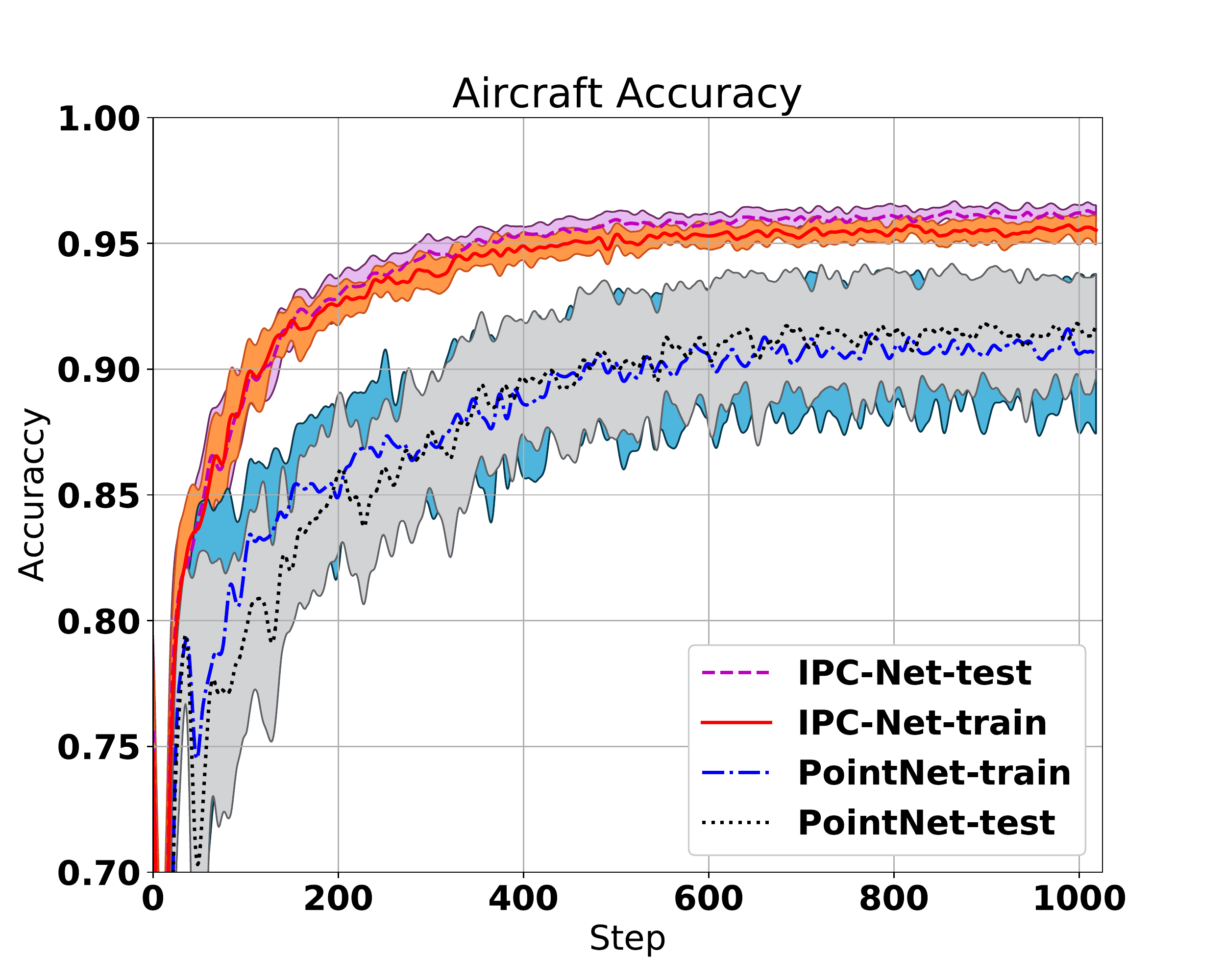}
          \caption{Aircraft}
          \label{fig:acc-aircraft}
      \end{subfigure}
      \begin{subfigure}{0.5\textwidth}
         \centering
          \includegraphics[ width=1\textwidth]{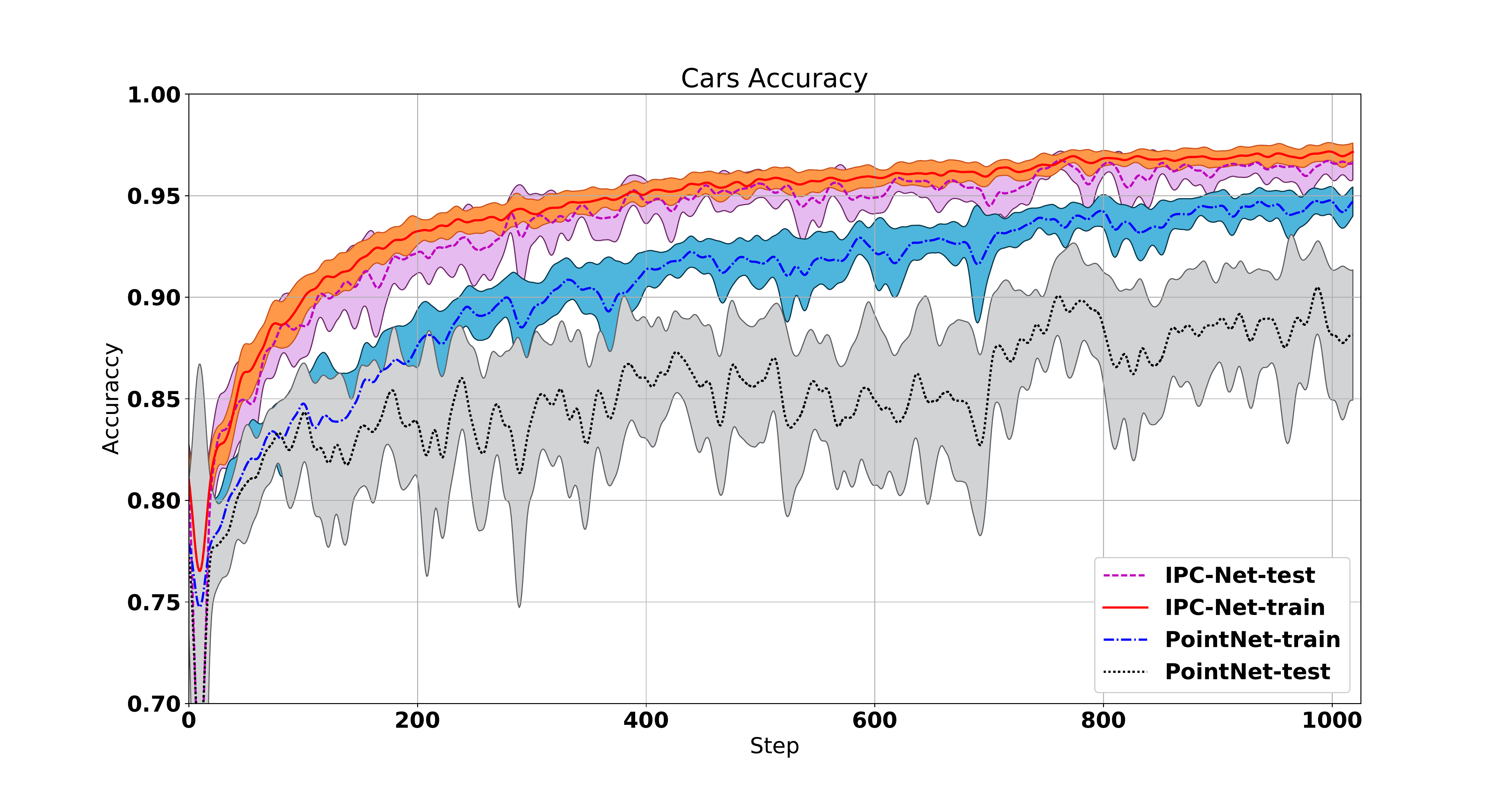}
          \caption{Cars}
          \label{fig:acc-cars}
      \end{subfigure}
     }

     \mbox{
     \begin{subfigure}{0.5\textwidth}
         \centering
          \includegraphics[width=1\textwidth]{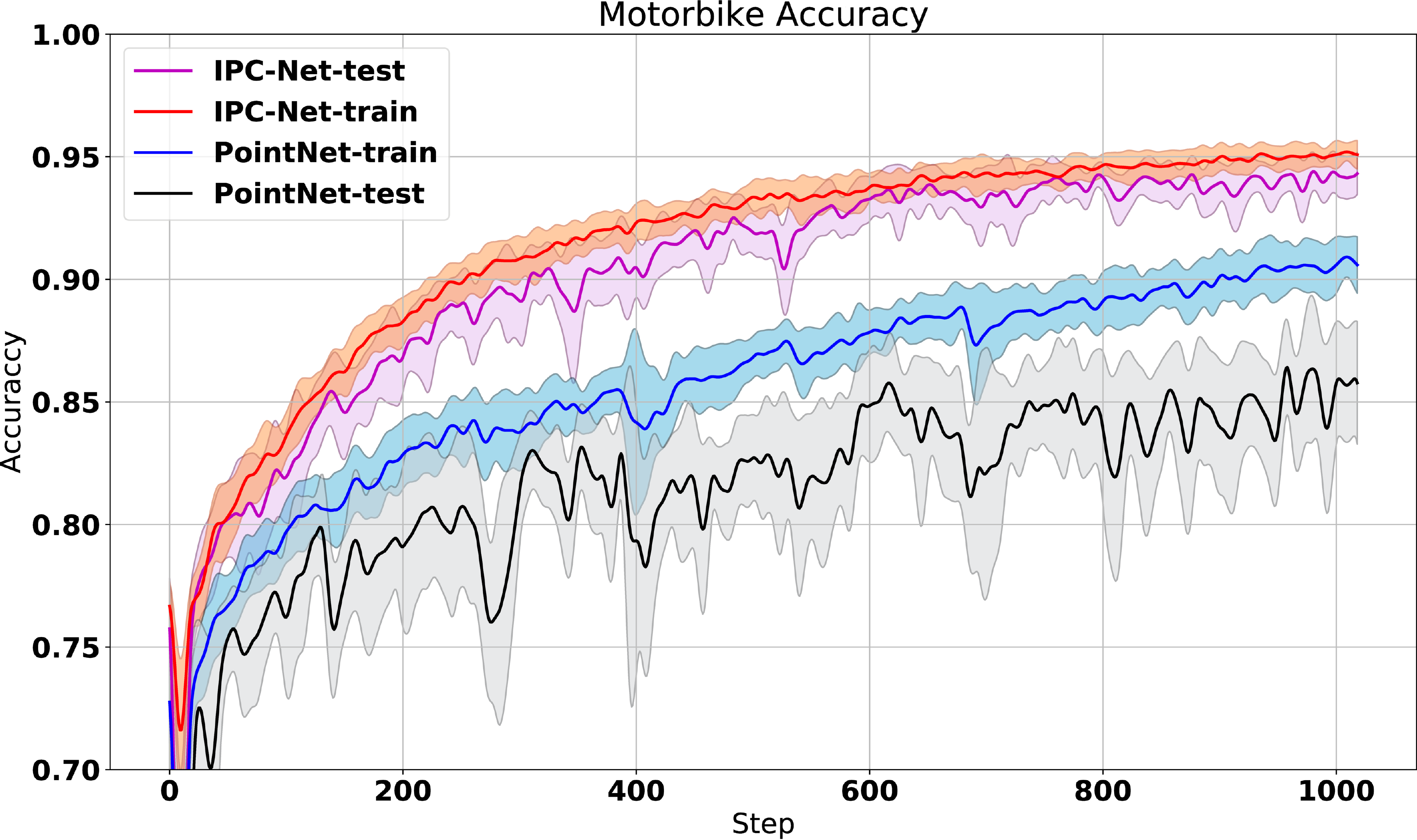}
          \caption{Motorbike}
          \label{fig:acc-bike}
     \end{subfigure}
     \begin{subfigure}{0.5\textwidth}
         \centering
          \includegraphics[width=1\textwidth]{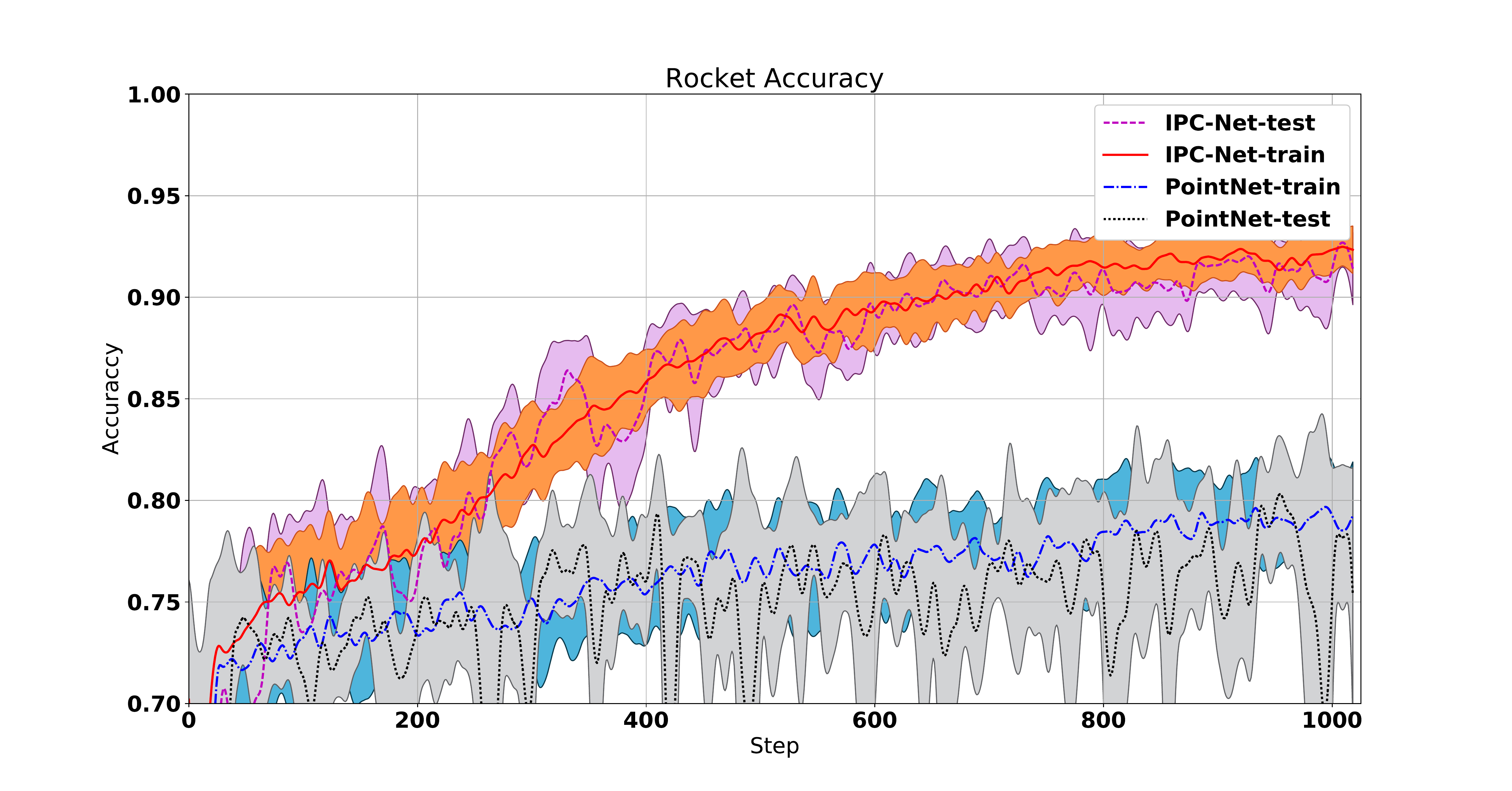}
          \caption{Rocket}
          \label{fig:acc-rocket}
      \end{subfigure}
      }
      \caption{This figure illustrates the mean and variance of the networks' accuracies over different runs. The full line and the dotted line represent the training and validation accuracy of the IPC-Net respectively. The symbol line and the dashed line represent the training and validation accuracy of the PointNet. One can observe that IPC-Net outperforms the accuracy of the PointNet. Additionally, our model contains less variance and learns considerably faster.}
      \label{fig:acc}
\end{figure*}
\begin{figure}%[!thbp]
    \centering
      \mbox{
      \begin{subfigure}{0.5\textwidth}
         \centering
          \includegraphics[width=0.9\textwidth,clip, trim=3cm 0cm 3cm 0cm]{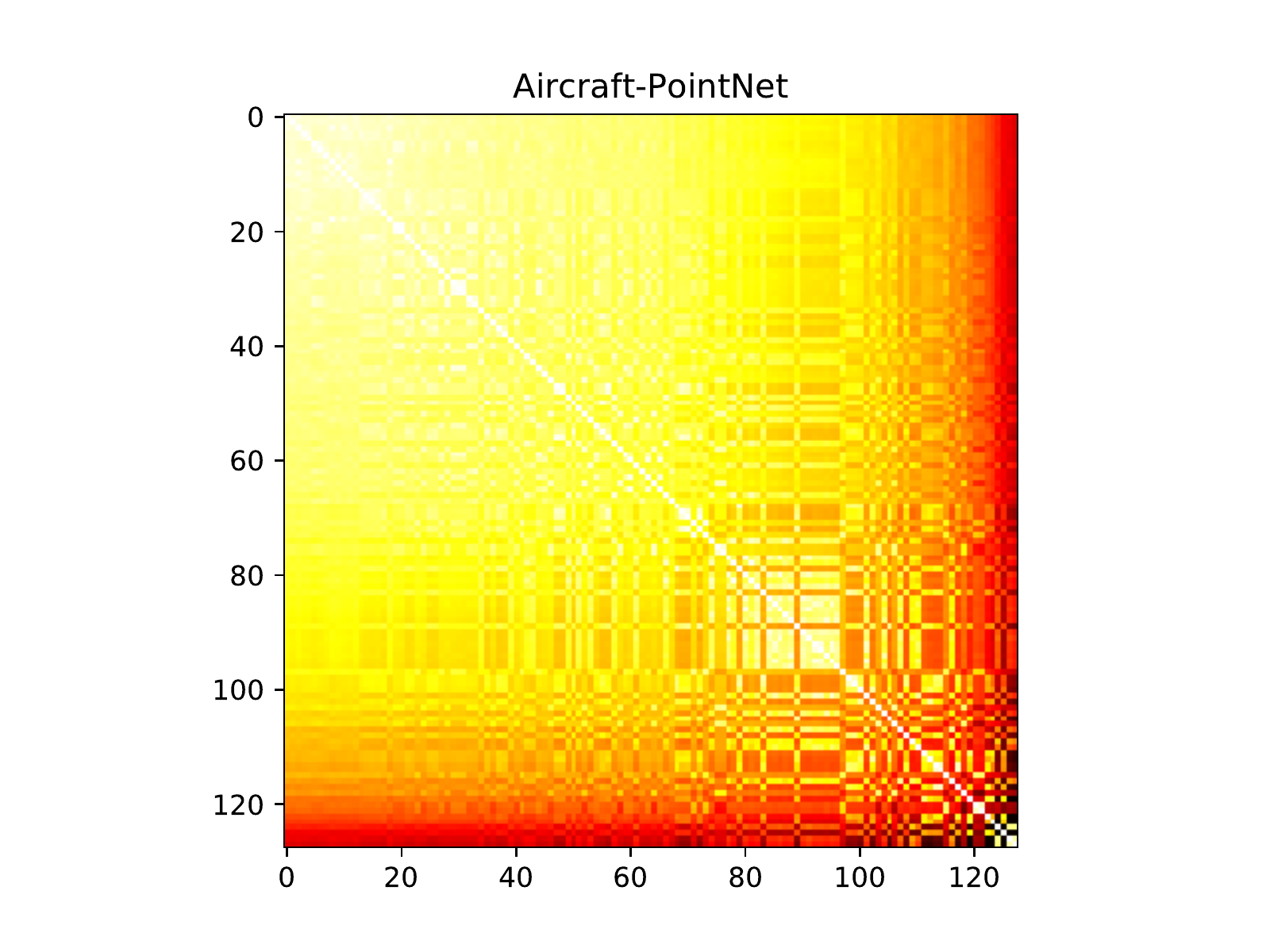}
          \caption{PointNet}
          \label{fig:norm-heat_kernels}
      \end{subfigure}
      \begin{subfigure}{0.5\textwidth}
        \centering
        \includegraphics[width=0.9\textwidth,clip, trim=3cm 0cm 3cm 0cm]{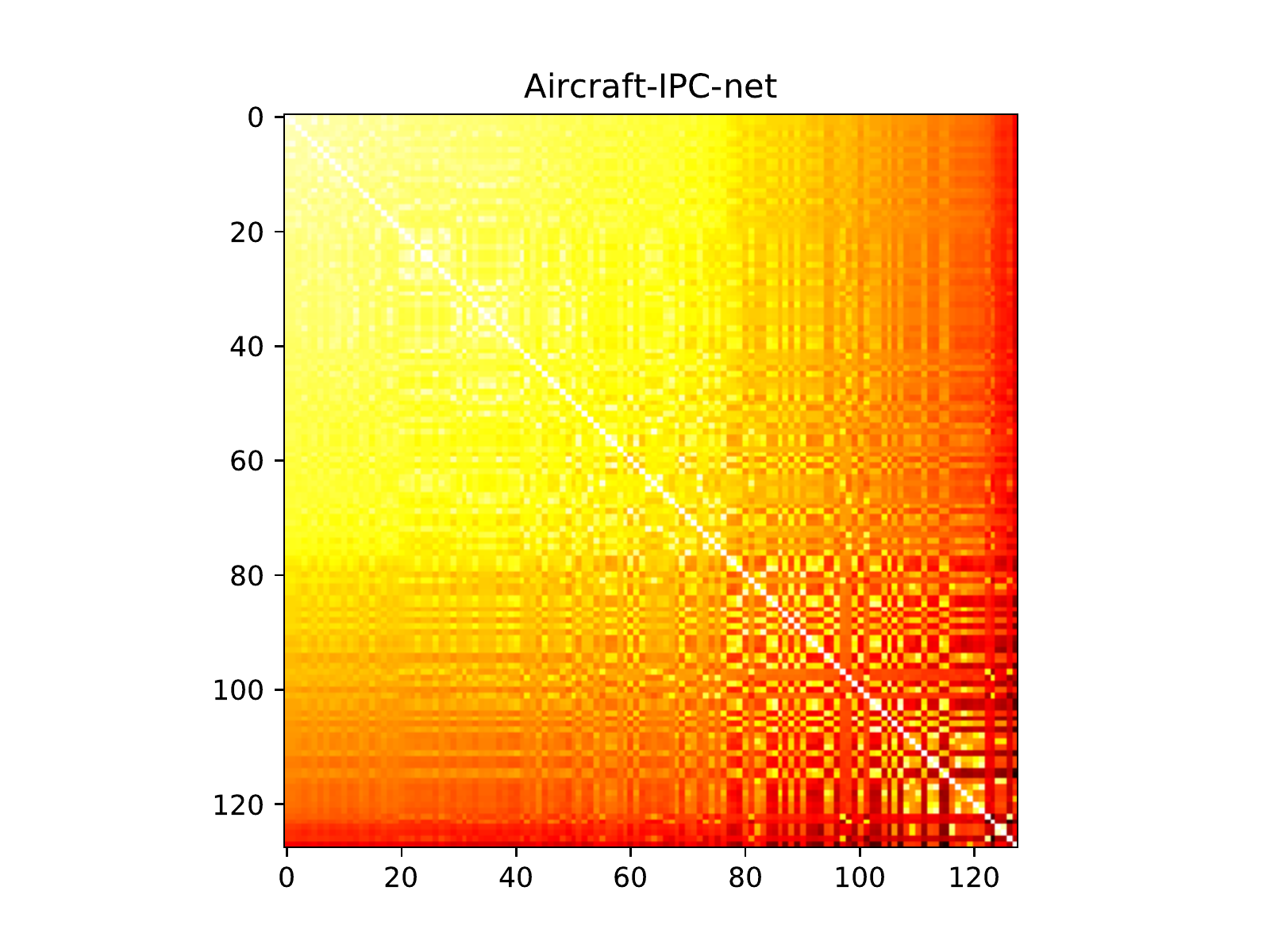}
          \caption{ICP-Net}
          \label{fig:conv-heat_kernels}
      \end{subfigure}
      }
%\end{figure}
%\begin{figure}%[!thbp]
    %\centering
      \mbox{
      \begin{subfigure}{0.5\textwidth}
         \centering
          \includegraphics[width=0.9\textwidth,clip, trim=3cm 0cm 3cm 0cm]{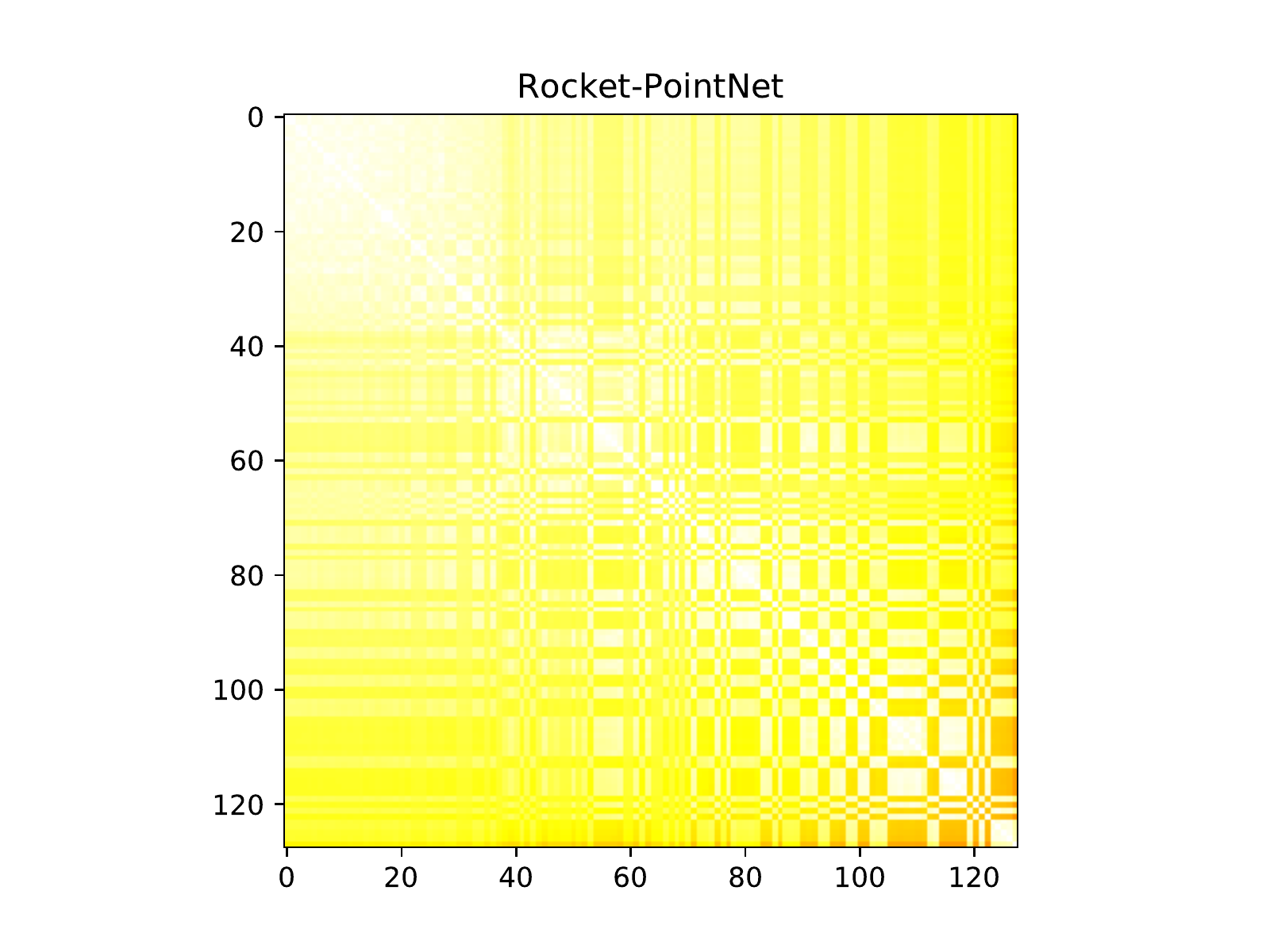}
          \caption{PointNet}
          \label{fig:rocket-norm-heat_kernels}
      \end{subfigure}
      \begin{subfigure}{0.5\textwidth}
        \centering
        \includegraphics[width=0.9\textwidth,clip, trim=3cm 0cm 3cm 0cm]{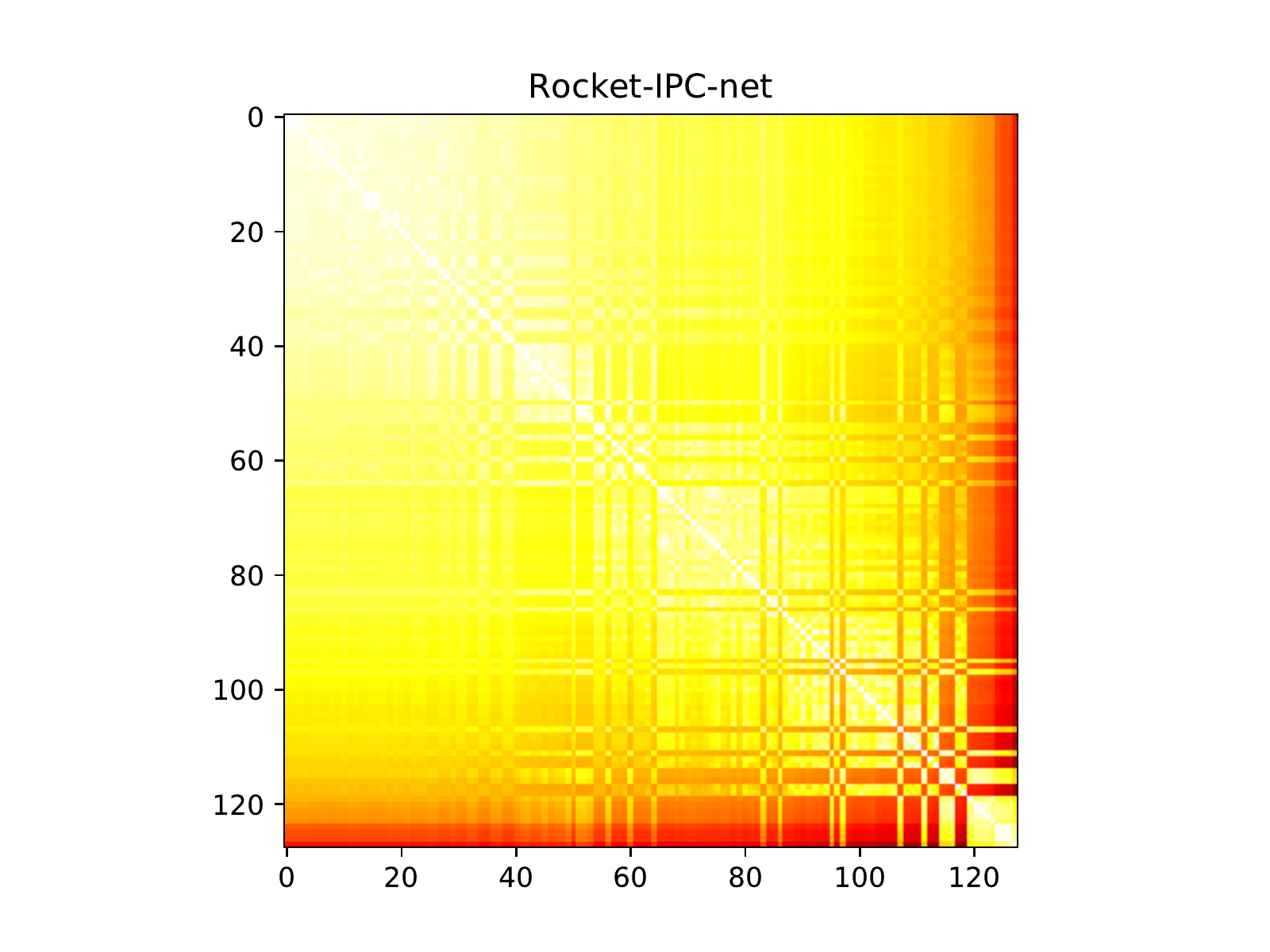}
          \caption{ICP-Net}
          \label{fig:rocket-conv-heat_kernels}
      \end{subfigure}
      }
      \caption{This figure illustrates the redundancy of the kernels by means of a distance heatmap. Each pixel $i,j$ symbolises the Euclidean distance between the feature vectors of kernels i and j . In order to make the redundancy on the kernels more explicit, the kernels  were ordered  based on the total distances between other kernels such that the more unique kernels appeared at the bottom right of the heat map.}
      %\caption{This figure illustrates the kernel redundancy of the rocket dataset. This figure explains the difference between the PointNet and the IPC-Net mean accuracy of figure \ref{fig:acc-rocket}   }
      \label{fig:heat-kernels}
      %\label{fig:rocket-heat-kernels}
\end{figure}

\begin{figure*}[!t]
    \centering
    \includegraphics[width=1.0\textwidth]{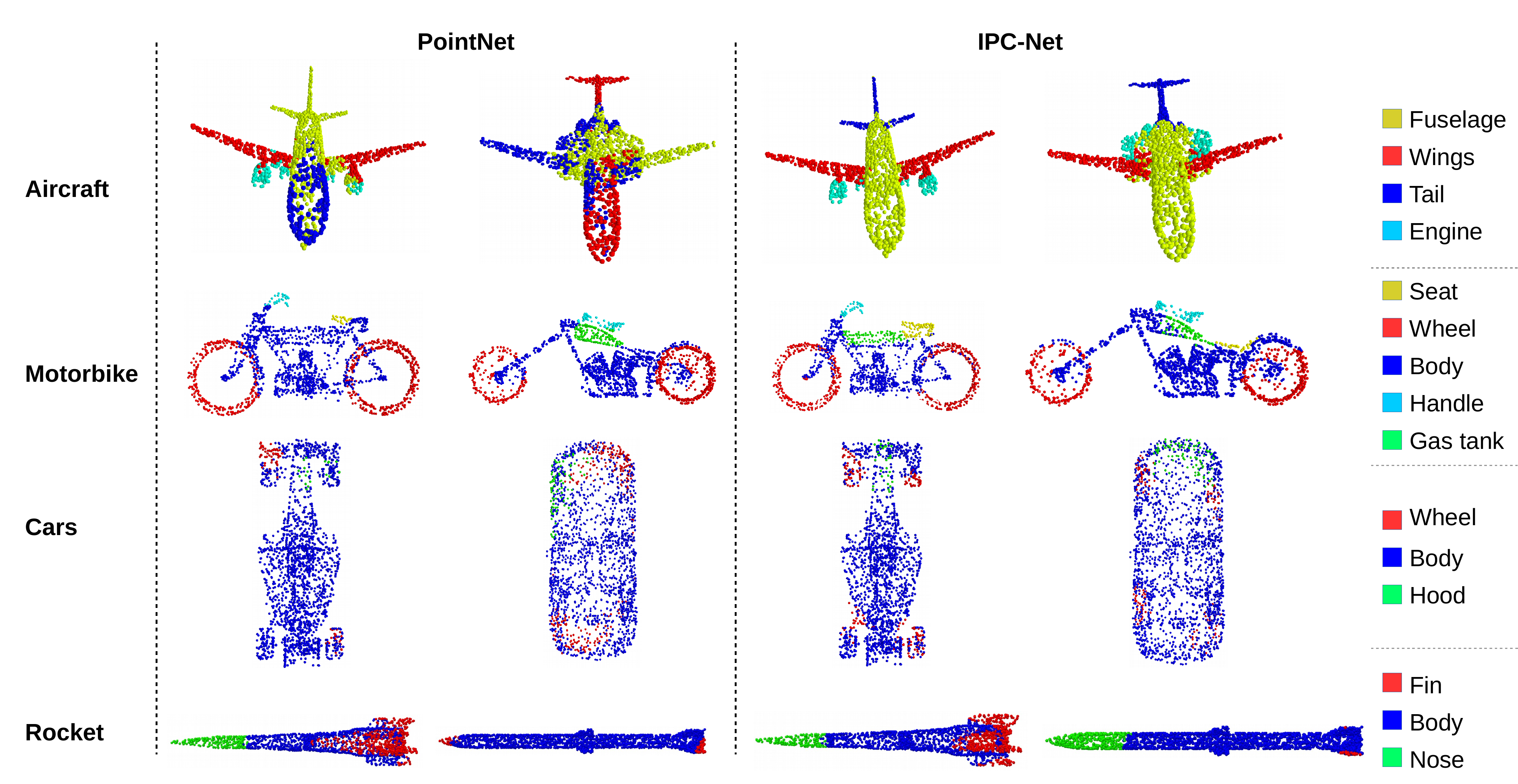}
    \caption{This figure illustrates the prediction of the Pointnet and the IPC-Net over 4 family of geometries that were sampled according to equation \ref{eq:sampling}. We can observe that in most cases the IPC-Net managed to achieved better segmentation than the point net as is reflected in figure \ref{fig:acc}.}
    \label{fig:geom-pred}
\end{figure*}

\subsection{Datasets}\label{sec:datasets}
Our convolutional model was trained on the segment-annotated family of the ShapeNet dataset \cite{chang2015}. This dataset comes in two flavours: the original dataset which contains roughly 51,300 unique 3D models categorised into different family groups, and the \emph{annotated-ShapeNet} \cite{yi2016scalable} which contains a labelled subset  of shapes from the original ShapeNet. The annotated-ShapeNet dataset contains 16,881 shapes from 16 different categories. In contrast to the original dataset, the annotated-ShapeNet dataset holds a point-cloud representation of the original shapes where the (ground truth) annotations are labelled for each point in the point-cloud. We selected the Aircraft category to perform the kernel analysis that was described in section \ref{sec:pointNetView}. In addition, we selected the \emph{Car}, \emph{Motorbike} and \emph{Rocket} categories to compare the segmentation accuracies between the PointNet and the IPC-Net. Each chosen dataset was randomly  partitioned into training and test sets and run over several trials to evaluate the robustness and generalisation accuracy of IPC-Net. Table \ref{tab:properties} illustrates an overview of the properties of each dataset. 

\begin{table}[htbp]
  \caption{Properties of the annotated-ShapeNet dataset}
  \begin{center}
    \renewcommand{\arraystretch}{1.2} 
      \begin{tabular}{|c|c|c|p{4cm}|}
      \hline
      \textbf{Shape}&\multicolumn{3}{|c|}{\textbf{Dataset partition}} \\
      \cline{2-4} 
      \textbf{Family} & \textbf{\textit{Train}} ($80\%$)& \textbf{\textit{Test}} ($20\%$)& \textbf{\textit{Labels}} \\
      \hline
      \textbf{Aircraft}& {3329}& {807}& [Body,Engine,Tail,Wing] \\
      \hline
      \textbf{Cars}& {400}& {100}&  [Body,Hood,Wheel]\\
      \hline
      \textbf{Motorbikes}& {269}& {67}&  [Body,Handle,Gas-tank,Seat,Wheel]\\
      \hline
      \textbf{Rockets}& {60}& {15}&  [Body,Nose,Fin]\\
      \hline
    \end{tabular}
    \label{tab:properties}
  \end{center}
\end{table}

Although the test set fully reflects the generalisation capacity of the models as it was not used to optimise the hyperparameters of the network, the model that yields the best accuracy in the test set was chosen to be the final model of the network. Therefore, we are indirectly introducing part of the test model into the learning phase. Consequently, the generalisation of the models was further compared visually by creating a validation set (see Section \ref{seg:sampling}). This was done to ensure that the models are not tested on geometries that were used in the training phase and to guarantee that the geometries are robust to a different point-sampling technique. 

\subsection{Sampling strategy}\label{seg:sampling}
The 3D shapes for the validation set were extracted from the original ShapeNet dataset as the annotated-ShapeNet dataset only holds a subset of this dataset. Due to the fact that the original shapes are not in a point-cloud representation, we sampled the triangular surface of the meshes to generate a point-cloud. Prior to sampling, we first normalise the 3D shapes $X$ according to the \emph{unit sphere normalisation} shown in the following formulas, as both models are not invariant to scale transformations.

\begin{equation}
    \label{eq:center}
     centre_{x,y,z} = \big[ \max_i X_{i,j} - \min_i X_{i,j}\big]^3_{j=1} \times 0.5 
\end{equation}
\begin{equation}
    X_{usphere} = \frac{X - 1^Tcentre_{x,y,z}}{ \max_i \norm{X_i - centre_{x,y,z}}}
    \label{eq:unit-norm}
\end{equation}

Equation \ref{eq:center} is responsible for centring the geometries to the origin $O$ in Euclidean space for each $[x,y,z]$ coordinates. Equation \ref{eq:unit-norm} normalises the centred geometries to a unit sphere. After normalising each shape, $N$ points were uniformly sampled from the surface of the triangulation of the 3D shape. A triangulation is a common mesh representation of 3D geometries that approximates the shape of the 3D geometry by a set of connected triangles. To ensure that the overall surface of the 3D shape is captured, the surface of a triangle was sampled proportional to the area times its equilaterality ratio. To calculate the probability that a triangle $t$ will be sampled, we used a binary search algorithm on the cumulative area and in the weighted ratio distribution of the edges \cite{Osada2002}. The coordinates of the  points at the surface of a triangle were calculated with the following equation:

\begin{equation}
  \mathbf{P} = (1 - \sqrt{r_1}) \vec{\mathbf{a}} + \sqrt{r_1}(1-r_2) \vec{\mathbf{b}}+\sqrt{r_1 r_2} \vec{\mathbf{c}}
  \label{eq:sampling}
\end{equation}

In this equation $[\vec{\mathbf{a}}, \vec{\mathbf{b}}, \vec{\mathbf{c}}]$ are the vertices of the triangle, $r_1$ is the percentage distance from vertex $\mathbf{A}$ to the opposite edge and $r_2$ is the percentage along the opposite edge \cite{yu2011three}.

\subsection{Kernel Analysis}
We selected a group of aircraft to analyse the kernels of the original model. Figure \ref{fig:norm-kernels} shows the kernel visualisation of two hidden layers in the PointNet network. Each visualisation was obtained by multiplying a \emph{feature kernel vector} with the point-cloud. This results into a sparse matrix were the non-zero values are the activations (important features) for that particular kernel. In these figures we can observe that several patterns of activations have been learned. These activations are a complex collection of approximated 2D planes that partition the 3D space. We visualised the kernels of the lower and higher layers as most kernels throughout the network yield highly similar features. In figure \ref{fig:norm-lkernels} we can observe that most partitions lean towards a more linear fragmentation (i.e., less complex features) compared to those in Figure \ref{fig:norm-hkernels}. This was expected as features that are closer to the higher layers will naturally contain more complex representations.

It is important to note that both figures are a 2D projection of the kernels. Therefore, there may be features that are not relevant for this 2D projection. In regards to feature activation, we note that in most cases the partitions in the feature space yield clusters of points that are close to each other in Euclidean space (e.g., points that belong only to a wing or to the fuselage of the plane). This inner-information that exists within local activations is exploited by our architecture as described in section \ref{sec:convpointNet}.

For every dataset in Table \ref{tab:properties}, we plotted the mean accuracy and the variance of the IPC-Net and PointNet. In Figure \ref{fig:acc} we perceive a noticeable improvement for the IPC-Net. The biggest improvement can be seen in the rocket's dataset where the gap accuracy between the two models is considerably bigger. Additionally, from this Figure, we observe that our model managed to learn the correct segmentations considerably faster and it is substantially  more consistent over the different runs. This performance improvement is due to the exploitation of neighbouring features that exist within the kernels of the PointNet. These new features allow the network to produce partitions that are more informative for the segmentation task. For example, based on the heatmap in Figure \ref{fig:heat-kernels} the PointNet architecture extracts insufficient information from the data to correctly guide the partitioning of the feature space. This statement is derived from the kernel redundancy (light colours) found in the last layers of the PointNet as shown Figure \ref{fig:norm-heat_kernels}. In contrast, due to the exploitation of neighbouring kernel information, the IPC-Net managed to extract a larger set of unique kernels reducing the amount of redundant features. This is reflected in Figure \ref{fig:conv-heat_kernels} where the red denotes the uniqueness of the features in the kernels. Furthermore, Figure \ref{fig:heat-kernels} provides a complementary heat map to the rocket Figure in \ref{fig:acc-rocket} which explains the large  difference in accuracy between the IPC-Net and the PointNet.  This Figure shows that the majority of kernels in PointNet are very similar (i.e, more redundant) which leads the model to suggest the wrong segmentations. In contrast, Figure \ref{fig:rocket-conv-heat_kernels} shows a large spectrum of unique features that aids the model to produce more desirable segmentations.

\subsection{Results}
\begin{table}[!htbp]
  \tiny
   \begin{center}
     \caption[IPC-Net results]{IPC-Net results for the ShapeNet dataset . The metric used is the mean IoU(\%) on the points.}
     \renewcommand{\arraystretch}{1.5}
         \begin{tabular}{@{\hskip 1pt}p{1.6cm}|@{\hskip 2pt}p{0.4cm}|@{\hskip 2pt}p{0.4cm}|@{\hskip 2pt}p{0.4cm}|@{\hskip 2pt}p{0.4cm}|@{\hskip 2pt}p{0.4cm}|@{\hskip 2pt}p{0.4cm}|@{\hskip 2pt}p{0.4cm}|@{\hskip 2pt}p{0.4cm}|@{\hskip 2pt}p{0.4cm}|@{\hskip 2pt}p{0.4cm}|@{\hskip 2pt}p{0.4cm}|@{\hskip 2pt}p{0.4cm}|@{\hskip 2pt}p{0.4cm}|@{\hskip 2pt}p{0.4cm}|@{\hskip 2pt}p{0.4cm}|@{\hskip 2pt}p{0.4cm}}
          &\hspace{5pt}\textbf{\rot{Aircraft}}&\textbf{\hspace{4pt}\rot{Bag}}&\textbf{\hspace{4pt}\rot{Cap}}&\textbf{\hspace{4pt}\rot{Car}}&\textbf{\hspace{4pt}\rot{Chair}}&\textbf{\hspace{4pt}\rot{Earphone}}&\textbf{\hspace{4pt}\rot{Guitar}}&\textbf{\hspace{4pt}\rot{Knife}}&\textbf{\hspace{4pt}\rot{Lamp}}&\textbf{\hspace{4pt}\rot{Laptop}}&\textbf{\hspace{4pt}\rot{Motorbike}}&\textbf{\hspace{4pt}\rot{Mug}}&\textbf{\hspace{4pt} \rot {Pistol}}&\textbf{\hspace{4pt}\rot{Rocket}}&\textbf{\hspace{4pt}\rot{Skateboard}}&\textbf{\hspace{4pt}\rot{Table}} \\\hline
          \textbf{\#Shapes}& 2690 & 76  & 55 &898  &3758  &69  &787  &392  &1547  &451  &202  &184  &283  &66  &152  &5271  \\\hline
          \textbf{\citea{wu2014interactive}}&63.2 & - & - & - & 73.5 & - &-  &-  &74.4  &-  &-  &-  & - & - & - &74.8  \\\hline
          \textbf{Kd-Tree \citea{Klokov2017}}&79.4 & 71.9 & 80.9 & 68.8 & 88.0 & 72.4 & 88.9  &86.4  &79.8  &94.9  &55.8  &86.5  & 79.3 & 50.4 & 71.1 & 80.2  \\\hline
          \textbf{3DCNN \citea{Qi2017}}&75.1 &72.8  &73.3  &70.0  &87.2  &63.5  &88.4  &79.6  &74.4  &93.9  &58.7  &91.8  &76.4  &51.2  &65.3  &77.1  \\\hline
          \textbf{Yi \citea{yi2016scalable}} &81.0  &78.4  &77.7  & 75.7&87.6  &61.9  &92.0 & 85.5 &82.5  & 95.7  &70.6  & 91.9 &85.9  &53.1  &69.8  &75.3  \\\hline
          \textbf{PointNet \citea{Qi2017}}&83.4 & 78.7 &82.5  & 74.9 &89.6  &73.0  &91.5  &85.9  & 80.8 & 95.3  & 65.2  &93.0 & 81.2  &57.9  & 72.8 &80.6  \\\hline
          \textbf{Yi \citea{yi2017syncspeccnn}}&81.6 &81.7 &81.9  &75.2  &90.2  &74.9  &93.0  &86.1  &84.7  &95.6  &66.7  &92.7  &81.6  &60.6 &82.9 &82.1  \\\hline
          \textbf{PointNet++ \citea{qi2017pointnet++}}&82.4 &79.0  &87.7  &77.3  &90.8  &71.8  &91.0  &85.9  &83.7  &95.3  &71.6  &94.1  &81.3  &58.7 &76.4 &82.6  \\\hline
             \textbf{IPC-Net}& \textbf{94.4} & \textbf{95.5}  & \textbf{95.9}  & \textbf{89.5}  & \textbf{95.2}  & \textbf{95.3} &\textbf{97.3}  &\textbf{96.6}  &\textbf{96.3}  &\textbf{98.2}  & \textbf{89.2}    &\textbf{94.8}  &\textbf{96.4}  & \textbf{86.2} &\textbf{95.3} & \textbf{96.2}  \\\hline
       \end{tabular}
     \label{tab:ipc-net}
   \end{center}
 \end{table}

\subsection{Visualisation results}
We visually evaluated the segmentation of 4 geometry families as shown in Figure \ref{fig:geom-pred}. This visualisation was done to evaluate the generalisation of the models by predicting  on geometries that were sampled with a different sampling technique as described in Section \ref{sec:datasets}. In this figure we can observe that our model accomplished a better generalisation accuracy compared to the PointNet model. For instance, the second aircraft prediction of Figure \ref{fig:geom-pred} shows that the IPC-Net kernels responsible for extracting circular information from engines, extrapolate this knowledge to a circular object found in the fuselage of the aircraft. This shows a clear example on how local neighbouring activations are essential to label other parts of geometries that share similar characteristics. Additionally, it also shows that the generalisation remains consistent on the rocket dataset.
%Regarding the geometries that were sampled according equation \ref{eq:sampling}, 

\section{Conclusion \& Discussion}

We propose an enhanced network architecture called IPC-Net that exploits the inner-information that exists within the kernel activations of the PointNet. A full kernel analysis was additionally provided which confirmed that the PointNet architecture disregards important information. We showed that the IPC-Net model is able to extract a more unique set of features which lead it to surpass the segmentation accuracy of the original architecture. This was clearly noticeable in every dataset where on average a large accuracy gap between the two models was observed. We additionally showed by means of heat map kernels the reason why the IPC-Net is more accurate and also learns considerably faster and more robustly across different family of geometries. 

\subsection{Discussion}

While our work brings a notable improvement, there are some aspects that could be enhanced in future research. For example, in the predictions of the car dataset,  we notice that the improvement over the original model is not as prominent as on the other datasets. Especially when predicting the \emph{hood} label on cars that had a symmetrical structure (i.e., both the front and the rear of the car are highly similar). This behaviour can be explained by the similarity of neighbouring activations found for both front and rear parts of the cars. We believe that the global features of the network do not sufficiently influence the neighbouring features of the network which  leads the model to a wrong segmentation prediction. This could be solved by increasing the number of samples in the dataset such that the global features become more prominent in the network. As a result, it will provide a better point of reference on where the neighbouring features are located in Euclidean space. Another solution is to increase the number of labels such that the global reference of the segmentations becomes clearer. A further analysis of this symmetric limitation needs to be investigated.    

Additionally, we found that in certain segmentations, an isolated cluster of misclassified segmentations can be observed. Similar to the symmetric limitation, we belief that these random misclassified clusters arise due to similar neighbouring characteristics that are found across the 3D shape.  As a result, if the global reference is not prominent, it will influence the network to perform a misclassification. This problem could potentially be improved by using Conditional Random fields \cite{lafferty2001conditional} which are popular in the field of image segmentation. This method influences the model to punish points that are comprised of different labels and are near each other. This technique could be used to smooth these random clusters that arise due to similar kernels.

%% file: pointlearning-arxiv.bbl
\begin{thebibliography}{29}
\providecommand{\natexlab}[1]{#1}
\providecommand{\url}[1]{\texttt{#1}}
\expandafter\ifx\csname urlstyle\endcsname\relax
  \providecommand{\doi}[1]{doi: #1}\else
  \providecommand{\doi}{doi: \begingroup \urlstyle{rm}\Url}\fi

\bibitem[Allen(1971)]{allen1971mean}
David~M Allen.
\newblock Mean square error of prediction as a criterion for selecting
  variables.
\newblock \emph{Technometrics}, 13\penalty0 (3):\penalty0 469--475, 1971.

\bibitem[Bottou(2010)]{bottou2010large}
L{\'e}on Bottou.
\newblock Large-scale machine learning with stochastic gradient descent.
\newblock In \emph{Proceedings of COMPSTAT'2010}, pages 177--186. Springer,
  2010.

\bibitem[Brock et~al.(2016)Brock, Lim, Ritchie, and Weston]{Brock2016}
Andrew Brock, Theodore Lim, James~M Ritchie, and Nick Weston.
\newblock Generative and discriminative voxel modeling with convolutional
  neural networks.
\newblock \emph{arXiv preprint arXiv:1608.04236}, 2016.

\bibitem[Chang et~al.(2015)Chang, Funkhouser, Guibas, Hanrahan, Huang, Li,
  Savarese, Savva, Song, Su, et~al.]{chang2015}
Angel~X Chang, Thomas Funkhouser, Leonidas Guibas, Pat Hanrahan, Qixing Huang,
  Zimo Li, Silvio Savarese, Manolis Savva, Shuran Song, Hao Su, et~al.
\newblock Shapenet: An information-rich 3d model repository.
\newblock \emph{arXiv preprint arXiv:1512.03012}, 2015.

\bibitem[Dieleman et~al.(2016)Dieleman, De~Fauw, and Kavukcuoglu]{Dieleman2016}
Sander Dieleman, Jeffrey De~Fauw, and Koray Kavukcuoglu.
\newblock Exploiting cyclic symmetry in convolutional neural networks.
\newblock \emph{arXiv preprint arXiv:1602.02660}, 2016.

\bibitem[Haenssle et~al.(2018)Haenssle, Fink, Schneiderbauer, Toberer, Buhl,
  Blum, Kalloo, Hassen, Thomas, Enk, and Uhlmann]{mdy166}
H~A Haenssle, C~Fink, R~Schneiderbauer, F~Toberer, T~Buhl, A~Blum, A~Kalloo,
  A~Ben~Hadj Hassen, L~Thomas, A~Enk, and L~Uhlmann.
\newblock Man against machine: diagnostic performance of a deep learning
  convolutional neural network for dermoscopic melanoma recognition in
  comparison to 58 dermatologists.
\newblock \emph{Annals of Oncology}, page mdy166, 2018.
\newblock \doi{10.1093/annonc/mdy166}.
\newblock URL \url{http://dx.doi.org/10.1093/annonc/mdy166}.

\bibitem[Jaderberg et~al.(2015)Jaderberg, Simonyan, Zisserman,
  et~al.]{jaderberg2015spatial}
Max Jaderberg, Karen Simonyan, Andrew Zisserman, et~al.
\newblock Spatial transformer networks.
\newblock In \emph{Advances in neural information processing systems}, pages
  2017--2025, 2015.

\bibitem[Kingma and Ba(2014)]{kingma2014adam}
Diederik~P Kingma and Jimmy Ba.
\newblock Adam: A method for stochastic optimization.
\newblock \emph{arXiv preprint arXiv:1412.6980}, 2014.

\bibitem[Klokov and Lempitsky(2017)]{Klokov2017}
Roman Klokov and Victor Lempitsky.
\newblock {Escape from Cells: Deep Kd-Networks for the Recognition of 3D Point
  Cloud Models}.
\newblock \emph{Proceedings of the IEEE International Conference on Computer
  Vision}, 2017-October:\penalty0 863--872, 2017.
\newblock ISSN 15505499.
\newblock \doi{10.1109/ICCV.2017.99}.

\bibitem[Krizhevsky et~al.(2012)Krizhevsky, Sutskever, and
  Hinton]{krizhevsky2012imagenet}
Alex Krizhevsky, Ilya Sutskever, and Geoffrey~E Hinton.
\newblock Imagenet classification with deep convolutional neural networks.
\newblock In \emph{Advances in neural information processing systems}, pages
  1097--1105, 2012.

\bibitem[Lafferty et~al.(2001)Lafferty, McCallum, and
  Pereira]{lafferty2001conditional}
John Lafferty, Andrew McCallum, and Fernando~CN Pereira.
\newblock Conditional random fields: Probabilistic models for segmenting and
  labeling sequence data.
\newblock 2001.

\bibitem[LeCun et~al.(2015)LeCun, Bengio, and Hinton]{lecun2015deep}
Yann LeCun, Yoshua Bengio, and Geoffrey Hinton.
\newblock Deep learning.
\newblock \emph{nature}, 521\penalty0 (7553):\penalty0 436, 2015.

\bibitem[M{\o}ller(1993)]{moller1993scaled}
Martin~Fodslette M{\o}ller.
\newblock A scaled conjugate gradient algorithm for fast supervised learning.
\newblock \emph{Neural networks}, 6\penalty0 (4):\penalty0 525--533, 1993.

\bibitem[Osada et~al.(2002)Osada, Funkhouser, Chazelle, and Dobkin]{Osada2002}
Robert Osada, Thomas Funkhouser, Bernard Chazelle, and David Dobkin.
\newblock Shape distributions.
\newblock \emph{ACM Transactions on Graphics (TOG)}, 21\penalty0 (4):\penalty0
  807--832, 2002.

\bibitem[Qi et~al.(2017{\natexlab{a}})Qi, Su, Mo, and Guibas]{Qi2017}
Charles~R Qi, Hao Su, Kaichun Mo, and Leonidas~J Guibas.
\newblock Pointnet: Deep learning on point sets for 3d classification and
  segmentation.
\newblock \emph{Proc. Computer Vision and Pattern Recognition (CVPR), IEEE},
  1\penalty0 (2):\penalty0 4, 2017{\natexlab{a}}.

\bibitem[Qi et~al.(2017{\natexlab{b}})Qi, Yi, Su, and Guibas]{qi2017pointnet++}
Charles~Ruizhongtai Qi, Li~Yi, Hao Su, and Leonidas~J Guibas.
\newblock Pointnet++: Deep hierarchical feature learning on point sets in a
  metric space.
\newblock In \emph{Advances in Neural Information Processing Systems}, pages
  5105--5114, 2017{\natexlab{b}}.

\bibitem[Russakovsky et~al.(2015)Russakovsky, Deng, Su, Krause, Satheesh, Ma,
  Huang, Karpathy, Khosla, Bernstein, et~al.]{russakovsky2015imagenet}
Olga Russakovsky, Jia Deng, Hao Su, Jonathan Krause, Sanjeev Satheesh, Sean Ma,
  Zhiheng Huang, Andrej Karpathy, Aditya Khosla, Michael Bernstein, et~al.
\newblock Imagenet large scale visual recognition challenge.
\newblock \emph{International Journal of Computer Vision}, 115\penalty0
  (3):\penalty0 211--252, 2015.

\bibitem[Sansoni et~al.(2009)Sansoni, Trebeschi, and Docchio]{sansoni2009state}
Giovanna Sansoni, Marco Trebeschi, and Franco Docchio.
\newblock State-of-the-art and applications of 3d imaging sensors in industry,
  cultural heritage, medicine, and criminal investigation.
\newblock \emph{Sensors}, 9\penalty0 (1):\penalty0 568--601, 2009.

\bibitem[Shore and Johnson(1980)]{shore1980axiomatic}
John Shore and Rodney Johnson.
\newblock Axiomatic derivation of the principle of maximum entropy and the
  principle of minimum cross-entropy.
\newblock \emph{IEEE Transactions on information theory}, 26\penalty0
  (1):\penalty0 26--37, 1980.

\bibitem[Su et~al.(2015)Su, Maji, Kalogerakis, and Learned-Miller]{Su2015}
Hang Su, Subhransu Maji, Evangelos Kalogerakis, and Erik Learned-Miller.
\newblock Multi-view convolutional neural networks for 3d shape recognition.
\newblock In \emph{Proceedings of the IEEE international conference on computer
  vision}, pages 945--953, 2015.

\bibitem[Theologou et~al.(2015)Theologou, Pratikakis, and
  Theoharis]{theologou2015comprehensive}
Panagiotis Theologou, Ioannis Pratikakis, and Theoharis Theoharis.
\newblock A comprehensive overview of methodologies and performance evaluation
  frameworks in 3d mesh segmentation.
\newblock \emph{Computer Vision and Image Understanding}, 135:\penalty0 49--82,
  2015.

\bibitem[Vinyals et~al.(2015)Vinyals, Bengio, and Kudlur]{vinyals2015order}
Oriol Vinyals, Samy Bengio, and Manjunath Kudlur.
\newblock Order matters: Sequence to sequence for sets.
\newblock \emph{arXiv preprint arXiv:1511.06391}, 2015.

\bibitem[Weisstein()]{WeissteinEW}
Eric~W. Weisstein.
\newblock Symmetric function. {From MathWorld---A Wolfram Web Resource}.
\newblock Last visited on 08/03/2018.

\bibitem[Wu et~al.(2014)Wu, Shou, Wang, and Liu]{wu2014interactive}
Zizhao Wu, Ruyang Shou, Yunhai Wang, and Xinguo Liu.
\newblock Interactive shape co-segmentation via label propagation.
\newblock \emph{Computers \& Graphics}, 38:\penalty0 248--254, 2014.

\bibitem[Yi et~al.(2016)Yi, Kim, Ceylan, Shen, Yan, Su, Lu, Huang, Sheffer,
  Guibas, et~al.]{yi2016scalable}
Li~Yi, Vladimir~G Kim, Duygu Ceylan, I~Shen, Mengyan Yan, Hao Su, Cewu Lu,
  Qixing Huang, Alla Sheffer, Leonidas Guibas, et~al.
\newblock A scalable active framework for region annotation in 3d shape
  collections.
\newblock \emph{ACM Transactions on Graphics (TOG)}, 35\penalty0 (6):\penalty0
  210, 2016.

\bibitem[Yi et~al.(2017)Yi, Su, Guo, and Guibas]{yi2017syncspeccnn}
Li~Yi, Hao Su, Xingwen Guo, and Leonidas~J Guibas.
\newblock Syncspeccnn: Synchronized spectral cnn for 3d shape segmentation.
\newblock In \emph{CVPR}, pages 6584--6592, 2017.

\bibitem[Yu et~al.(2011)Yu, Lu, Luo, and Wang]{yu2011three}
Faxin Yu, Zheming Lu, Hao Luo, and Pinghui Wang.
\newblock \emph{Three-dimensional model analysis and processing}.
\newblock Springer Science \& Business Media, 2011.

\bibitem[Zaheer et~al.(2017)Zaheer, Kottur, Ravanbakhsh, Poczos, Salakhutdinov,
  and Smola]{zaheer2017deep}
Manzil Zaheer, Satwik Kottur, Siamak Ravanbakhsh, Barnabas Poczos, Ruslan~R
  Salakhutdinov, and Alexander~J Smola.
\newblock Deep sets.
\newblock In \emph{Advances in Neural Information Processing Systems}, pages
  3394--3404, 2017.

\bibitem[Zeiler and Fergus(2014)]{zeiler2014visualizing}
Matthew~D Zeiler and Rob Fergus.
\newblock Visualizing and understanding convolutional networks.
\newblock In \emph{European conference on computer vision}, pages 818--833.
  Springer, 2014.

\end{thebibliography}
